\documentclass{article} 
\usepackage{iclr2017_workshop,times}
\usepackage{hyperref}
\usepackage{url}

\usepackage{graphics}
\usepackage{amsmath}
\usepackage{graphicx}
\usepackage{float}
\usepackage[caption = false]{subfig}

\usepackage{longtable}

\usepackage{booktabs}
\usepackage{multirow}
\usepackage{siunitx}

\title{Knowledge distillation using unlabeled mismatched images }

\author{Mandar Kulkarni(*), Kalpesh Patil(**), Shirish Karande(*)\\
TCS Innovation Labs, Pune, India (*), IIT Bombay, Mumbai, India(**)\\
\texttt{\{mandar.kulkarni3,shirish.karande\}@tcs.com}, \texttt{kalpeshpatil@iitb.ac.in}\\
}

\begin{document}

\maketitle

\vspace{-0.5cm}
\begin{abstract}
Current approaches for Knowledge Distillation (KD)  either directly use training data or sample from the training data distribution. In this paper, we demonstrate effectiveness of 'mismatched' unlabeled stimulus to perform KD for image classification networks. For illustration, we consider scenarios where this is a complete absence of training data, or mismatched stimulus has to be used for augmenting a small amount of training data. We demonstrate that stimulus complexity is a key factor for distillation's good performance.  Our examples include use of various datasets for stimulating MNIST and CIFAR teachers.

\end{abstract}


\section{Introduction and Related works}

Knowledge Distillation (KD) is the process of transferring the generalization ability of a teacher model (usually large neural network) to a student model (usually a small neural network). 
\cite{hinton2015distilling} demonstrated that the student network can be trained using an input data with combination of hard labels as well as soft labels. The hard labels are the ground truth labels (one-hot vectors) available for the training data while the soft labels are the output of the teacher network on the input data. 
Most of the distillation approaches either directly use training data or device a strategy to learn the training data distribution and then sample from it.
However, the assumption of availability of labeled training data may not always hold true, due to various reasons. 





In this paper, we investigate an effectiveness of 'mismatched' unlabeled stimulus for KD when the training data of the teacher is not available. 
Specifically, for the MNIST teacher, we utilize mismatched stimulus such as CIFAR(\cite{krizhevsky2009learning}), STL(\cite{coates2010analysis}), Shape(\cite{bengio2009curriculum}) and Noise. For a CIFAR teacher, we use stimulus such as 120k Tiny Imagenet(\cite{torralba200880},\cite{tiny120k}), MNIST (\cite{lecun1998gradient}), Shape, SVHN (\cite{netzer2011reading}), DTD-Texture (\cite{cimpoi14describing}). We observe that for CNN architectures these stimuli provide a surprisingly efficient distillation to student networks.
We study an effect of complexity of the stimulus dataset on the distillation performance by using stimulus of varied complexity. Experimental results clearly demonstrate that the complexity of stimulus plays an important role in distillation, where more complex dataset appear to give better generalization performance. We also consider a scenario where a small labeled training set is available. In such cases, unlabeled stimulus can be very effective for data augmentation.

\section{Preliminaries}

\subsection{Methodology}
The training objective of the distillation process \cite{romero2014fitnets} can be written as follows
\begin{eqnarray}\label{eq:a1}
L(W_S) = H(P_T, P_S) + \beta  H(y_{true}, P_S) 
\end{eqnarray}
where $W_S$ indicates weights of the student network, $H$ indicates the cross entropy and $\beta$ is the relative weights of two terms. The second term in the equation corresponds to a traditional cross entropy loss between output of a student network and labels ($y_{true}$). 
Let $D$ denotes the data used for distillation.
$P_T$ is the posterior output of the teacher network on $D$ while $P_S$ is the posterior output of the student. The first term in the equation attempts to make the posterior of the student similar to that of teacher for the input data $D$. If the training data is not available, second term in the Eq. \ref{eq:a1} cannot be used. A student is trained to optimize only the first term in Eq. \ref{eq:a1}. 
In case of availability of a small labeled training set, we use combination of this labeled set and unlabeled stimulus to train a student. For the unlabeled stimulus, we assume an uniform distribution over all classes while using it in the second term of Eq. \ref{eq:a1}. 

\subsection{Description of datasets and networks}
We use teachers trained on two well known image classification datasets, MNIST and CIFAR-10. 

\subsubsection{MNIST}
A CNN is trained on the dataset which has approx. 478k params. We experiment with two student architectures. In the first experiment, we use a student with the same architecture as the teacher. 
In the second experiment, we attempt to distill the teacher network into a relatively smaller student CNN.  The student network has approx. 35\% less parameters than the teacher. The details of the teacher and student architectures are provided in appendix.
For MNIST teacher, we use mismatched stimulus such as CIFAR-10 , STL-10, Shape dataset and uniform random noise [-0.3,0.7]. In case of CIFAR and STL data, images are converted to grayscale and resized appropriately. 

\subsubsection{CIFAR}
We train a 12 layer CNN on the CIFAR dataset which has approx.3M params.
Here as well, we experiment with two student architectures, one which is same as the teacher and other smaller CNN which has approx. 10 times less parameters than the teacher. 
For CIFAR teacher, we use mismatched stimulus such as MNIST, SVHN, Shape, Texture, uniform noise and a slightly similar dataset, 120k TinyImagenet. 
As pointed out by the reviewer, a subset of 80M unlabeled TinyImagenet is been used for CIFAR distillation \cite{ba2014deep}.
However, we demonstrate the result on 120k TinyImagenet which contain labeled examples belonging to 200 classes.   
We observe that there is no significant overlap between the classes of CIFAR and 120k TinyImagenet. 



\section{Experimental results}


\subsection{Performance of mismatched stimulus and effect of its complexity}

\subsubsection{MNIST}
\begin{figure} [!t]
\flushleft
\hspace*{-2cm}\begin{tabular}{c c c c}
\subfloat[]{\includegraphics[height = 100pt,width = 120pt]{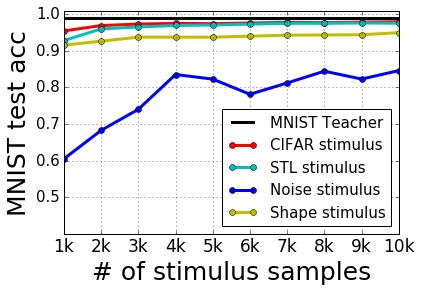}}
\subfloat[]{\includegraphics[height = 100pt,width = 120pt]{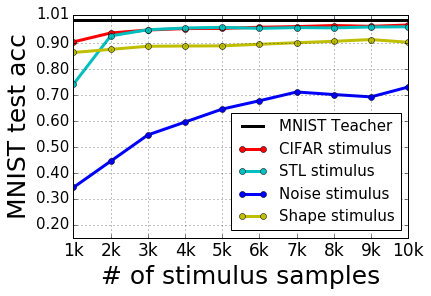}} 

\subfloat[]{\includegraphics[height = 100pt,width = 120pt]{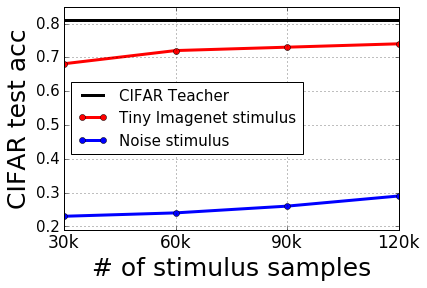}}
\subfloat[]{\includegraphics[height = 100pt,width = 120pt]{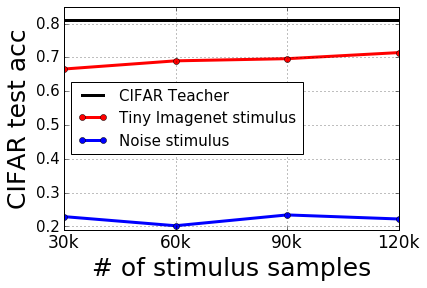}}

\end{tabular}
\caption{\label{fig:inpt2} Distillation result with MNIST and CIFAR teachers.(a) MNIST test accuracy where teacher and student architectures (CNNs) are identical, (b) MNIST test accuracy where student is smaller than the teacher (CNNs), (c) CIFAR test accuracy when student is same as teacher, (d) CIFAR test accuracy when student is 10 times smaller than teacher.  }
\end{figure}

Fig. \ref{fig:inpt2}(a) shows the performance of mismatched stimulus on the student network which has same architecture as the teacher. The teacher accuracy on the test is 99.1\% (90 errors). The best accuracy obtained with CIFAR, STL and random stimulus is 98.1\% (190 test errors), 97.7\% (228 test errors) and 84.5\% respectively. The result seems interesting because, though MNIST is a digit dataset, a mismatched object dataset CIFAR works very well as the stimulus.
 Fig. \ref{fig:inpt2}(b) shows the test accuracy performance for a smaller student network. 

Shape dataset was previously used for demonstrating an effectiveness of Curriculum learning \cite{bengio2009curriculum}.
The dataset consist of 10k examples of simple shape images. Due to small variability, the dataset is simpler than CIFAR or STL.
From the plots in Fig. \ref{fig:inpt2}(a)(b), it can be seen that Shape stimulus performs inferior to CIFAR and STL.

For the sake of completeness, we also performed experiment with DNN teacher-student for MNIST. The details of the experiment are given in Appendix.



\vspace{-0.3cm} 
\subsubsection{CIFAR}

Fig. \ref{fig:inpt2}(c) shows the result when teacher and student are identical. Teacher accuracy is 81.1\% while the best accuracy with 120k TinyImagenet stimulus was 74\%. Fig. \ref{fig:inpt2}(d) shows the performance with a (10 times) smaller student CNN. The maximum accuracy here with TinyImagenet is 71.4\%. 

For CIFAR teacher, we additionally perform experiment with MNIST, Shape, SVHN, Texture datasets on the smaller student CNN. We use 5k samples from each dataset as the stimulus.
Table \ref{table:cifar1} shows the result of the experiment.
The datasets are of varied 'complexity' (variations). Visually, the order of complexity is MNIST $<$ Shape $<$ SVHN $<$ Texture $<$ Tiny Imagenet.  
A trend similar to MNIST is observed where a more complex dataset performs better than the relatively less complex stimulus. We explored one of the quantification approach for complexity and results are reported in Appendix.

              
                 

\begin{table}[h]
  \small
  \hspace*{1cm}\begin{tabular}{lllllll}
    \toprule
    \multirow{2}{*}{} & 
              
                 
       {Noise} & {MNIST} & {Shape} & {SVHN} & {Texture} & {TinyImagenet}\\
      \midrule 
    CIFAR Test acc & 0.125  & 0.161 & 0.228 & 0.304 & 0.371 & 0.429\\
    \bottomrule
  \end{tabular}
  \caption{\label{table:cifar1} CIFAR test acc with 5k samples from different stimuli. }
\end{table}

\vspace{-0.3cm}
\subsection{Unlabeled stimulus for data augmentation}
Though we have shown results under the assumption of no training data, mismatched stimulus can also be effective for data augmentation. To validate this, we performed following experiments. For the MNIST teacher, we used 500 labeled samples from MNIST and (optionally) augmented it with 3k unlabeled samples from various stimuli. 
Results are given in the Table \ref{table:mnist}. 

\begin{table}[!h]
  \small
  \hspace*{2cm}\begin{tabular}{lllll}
    \toprule
    \multirow{2}{*}{} & 
     
                 
      {No augmentation} & {Noise(3k)} & {CIFAR(3k)} & {Shape(3k)} \\ 
      \midrule 
    MNIST Test acc & 0.955 & 0.956& 0.972  & 0.973 \\
    \bottomrule
  \end{tabular}
  \caption{\label{table:mnist} MNIST teacher data augmentation results. }
\end{table} 

For CIFAR teacher, we used 5k labeled samples from CIFAR dataset and (optionally) augmented it with 5k samples from various stimuli. The Table \ref{table:cifar} shows the results.
It can be seen that, in both the cases data augmentation helps the student to generalize better. 

\begin{table}[!h]
  \small
  \begin{tabular}{llllllll}
    \toprule
    \multirow{2}{*}{} & 
              
                 
      {No augmentation} & {Noise} & {MNIST} & {Shape} & {SVHN} & {Texture} & {TinyImagenet}\\
      \midrule 
    CIFAR Test acc  & 0.548 & 0.583  & 0.594 & 0.593 & 0.586 & 0.632 & 0.634  \\
    \bottomrule
  \end{tabular}
  \caption{\label{table:cifar} CIFAR test acc with data augmentation. }
\end{table}

\vspace{-0.3cm}
\section{Discussion and Conclusion}
As mentioned in Buciluǎ et. al. \cite{buciluǎ2006model}, though collecting synthetic stimulus is easy for images, it is crucial that the data should match the training data distribution. 
If the training data is not available, mismatched images surprisingly,  turn out to be a good stimulus. Experimental results demonstrate that the complexity of the stimulus plays a major role and a more complex dataset provides better performance. We explored a quantification approach for dataset complexity. When a small training set is available, an unlabeled stimulus is also effective for data augmentation . 


\bibliography{iclr2017_workshop}

\begin{thebibliography}{14}
\providecommand{\natexlab}[1]{#1}
\providecommand{\url}[1]{\texttt{#1}}
\expandafter\ifx\csname urlstyle\endcsname\relax
  \providecommand{\doi}[1]{doi: #1}\else
  \providecommand{\doi}{doi: \begingroup \urlstyle{rm}\Url}\fi

\bibitem[tin()]{tiny120k}
Tiny imagenet 120k: http://cs231n.stanford.edu/tiny-imagenet-100-a.zip,
  http://cs231n.stanford.edu/tiny-imagenet-100-b.zip.

\bibitem[Ba \& Caruana(2014)Ba and Caruana]{ba2014deep}
Jimmy Ba and Rich Caruana.
\newblock Do deep nets really need to be deep?
\newblock In \emph{Advances in neural information processing systems}, pp.\
  2654--2662, 2014.

\bibitem[Bengio et~al.(2009)Bengio, Louradour, Collobert, and
  Weston]{bengio2009curriculum}
Yoshua Bengio, J{\'e}r{\^o}me Louradour, Ronan Collobert, and Jason Weston.
\newblock Curriculum learning.
\newblock In \emph{Proceedings of the 26th annual international conference on
  machine learning}, pp.\  41--48. ACM, 2009.

\bibitem[Buciluǎ et~al.(2006)Buciluǎ, Caruana, and
  Niculescu-Mizil]{buciluǎ2006model}
Cristian Buciluǎ, Rich Caruana, and Alexandru Niculescu-Mizil.
\newblock Model compression.
\newblock In \emph{Proceedings of the 12th ACM SIGKDD international conference
  on Knowledge discovery and data mining}, pp.\  535--541. ACM, 2006.

\bibitem[Cimpoi et~al.(2014)Cimpoi, Maji, Kokkinos, Mohamed, , and
  Vedaldi]{cimpoi14describing}
M.~Cimpoi, S.~Maji, I.~Kokkinos, S.~Mohamed, , and A.~Vedaldi.
\newblock Describing textures in the wild.
\newblock In \emph{Proceedings of the {IEEE} Conf. on Computer Vision and
  Pattern Recognition ({CVPR})}, 2014.

\bibitem[Coates et~al.(2010)Coates, Lee, and Ng]{coates2010analysis}
Adam Coates, Honglak Lee, and Andrew~Y Ng.
\newblock An analysis of single-layer networks in unsupervised feature
  learning.
\newblock \emph{Ann Arbor}, 1001\penalty0 (48109):\penalty0 2, 2010.

\bibitem[Hinton et~al.(2015)Hinton, Vinyals, and Dean]{hinton2015distilling}
Geoffrey Hinton, Oriol Vinyals, and Jeff Dean.
\newblock Distilling the knowledge in a neural network.
\newblock \emph{arXiv preprint arXiv:1503.02531}, 2015.

\bibitem[Krizhevsky \& Hinton(2009)Krizhevsky and
  Hinton]{krizhevsky2009learning}
Alex Krizhevsky and Geoffrey Hinton.
\newblock Learning multiple layers of features from tiny images.
\newblock 2009.

\bibitem[LeCun et~al.(1998)LeCun, Bottou, Bengio, and
  Haffner]{lecun1998gradient}
Yann LeCun, L{\'e}on Bottou, Yoshua Bengio, and Patrick Haffner.
\newblock Gradient-based learning applied to document recognition.
\newblock \emph{Proceedings of the IEEE}, 86\penalty0 (11):\penalty0
  2278--2324, 1998.

\bibitem[Netzer et~al.(2011)Netzer, Wang, Coates, Bissacco, Wu, and
  Ng]{netzer2011reading}
Yuval Netzer, Tao Wang, Adam Coates, Alessandro Bissacco, Bo~Wu, and Andrew~Y
  Ng.
\newblock Reading digits in natural images with unsupervised feature learning.
\newblock In \emph{NIPS workshop on deep learning and unsupervised feature
  learning}, volume 2011, pp.\ ~5, 2011.

\bibitem[Papamakarios(2015)]{papamakarios2015distilling}
George Papamakarios.
\newblock Distilling model knowledge.
\newblock \emph{arXiv preprint arXiv:1510.02437}, 2015.

\bibitem[Romero et~al.(2014)Romero, Ballas, Kahou, Chassang, Gatta, and
  Bengio]{romero2014fitnets}
Adriana Romero, Nicolas Ballas, Samira~Ebrahimi Kahou, Antoine Chassang, Carlo
  Gatta, and Yoshua Bengio.
\newblock Fitnets: Hints for thin deep nets.
\newblock \emph{arXiv preprint arXiv:1412.6550}, 2014.

\bibitem[Torralba et~al.(2008)Torralba, Fergus, and Freeman]{torralba200880}
Antonio Torralba, Rob Fergus, and William~T Freeman.
\newblock 80 million tiny images: A large data set for nonparametric object and
  scene recognition.
\newblock \emph{IEEE transactions on pattern analysis and machine
  intelligence}, 30\penalty0 (11):\penalty0 1958--1970, 2008.

\bibitem[Yosinski et~al.(2014)Yosinski, Clune, Bengio, and
  Lipson]{yosinski2014transferable}
Jason Yosinski, Jeff Clune, Yoshua Bengio, and Hod Lipson.
\newblock How transferable are features in deep neural networks?
\newblock In \emph{Advances in neural information processing systems}, pp.\
  3320--3328, 2014.

\end{thebibliography}
\bibliographystyle{iclr2017_workshop}

\section{Appendix}

\subsection{Details of teacher and student architectures}
\subsubsection{MNIST}
The architecture of the teacher network is [Conv1(32,5,5)-MaxPool(2)-Conv2(64,5,5)-MaxPool(2)-FC(128)-Softmax(10)]. The architecture of  smaller student network is [Conv1(16,5,5)-Conv2(16,5,5)-Conv3(16,5,5)-MaxPool(2)-Conv4(16,5,5)-MaxPool(2)-FC(128)-Softmax(10)].

\subsubsection{CIFAR}
The architecture of the teacher network is [Conv1(32,3,3)-
Conv2(32,3,3)-MaxPool(2)-Conv3(64,3,3)-
Conv4(64,3,3)-
MaxPool(2)-Conv5(128,3,3)-
Conv6(128,3,3)-MaxPool(2)-FC(1024)-FC(512)-Softmax(10)].
The architecture of  smaller student network is [Conv1(32,5,5)-
Conv2(32,5,5)-MaxPool(2)-Conv3(32,5,5)-
Conv4(32,5,5)-
MaxPool(2)-Conv5(32,5,5)-
Conv6(32,5,5)-MaxPool(2)-
Conv7(32,3,3)-
FC(1000)-Softmax(10)].

\subsection{Result of DNN teacher-student for MNIST teacher}
\begin{figure} [!h]
\centering
\begin{tabular}{c}
\includegraphics[height = 120pt,width = 170pt]{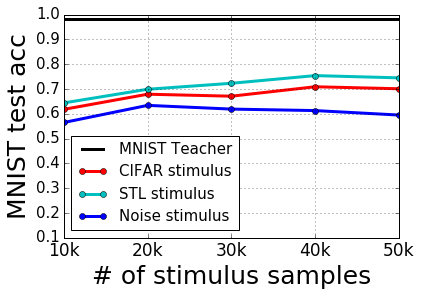}\\
\end{tabular}
\caption{\label{fig:sv1} DNN teacher-student for MNIST.}
\end{figure}

We experimented with DNN teacher-DNN student scenario similar to \cite{papamakarios2015distilling}. We train a single DNN teacher on MNIST data [784 - Dense(500) - Dense(300) - Softmax(10)]. The student is a DNN with only  10\% hidden nodes as compared to teacher. We perform distillation using CIFAR, STL and a normal Gaussian noise stimulus. Fig. \ref{fig:sv1} shows the result of the experiment. 

Note that, though natural image stimulus works better than noise, the difference in the performance
is not significant in case of DNN. It is known that the initial layers of CNN learn generic features
such as edges, blobs \cite{yosinski2014transferable}. Since such features are easily found in natural images as
compared to noise images, natural images turn out to be a be a better stimulus than noise. However,
since DNN employs full connection between hidden layers, its hidden nodes get activated even with random noise and hence noise may be performing well for DNN teacher as reported in \cite{papamakarios2015distilling}.

\subsection{Plot of cross-entropy loss and test accuracy}
Our objective function is to minimize the cross entropy loss between soft targets of the teacher and the student on the unlabeled stimulus. We terminate the iterations when the cross entropy loss cease to change. To visualize possibility of overfitting (if any), we plotted the cross entropy loss and the test accuracy for two cases: MNIST teacher using 1k CIFAR stimulus and CIFAR teacher using 1k Texture stimulus. The plots is shown below. 

\begin{figure} [!h]
\centering
\begin{tabular}{c c}
\includegraphics[height = 120pt,width = 170pt]{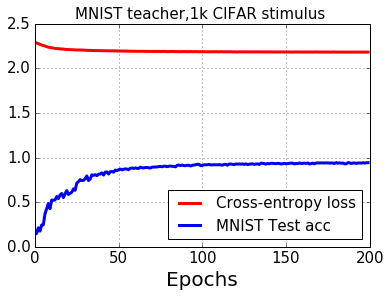}&
\includegraphics[height = 120pt,width = 170pt]{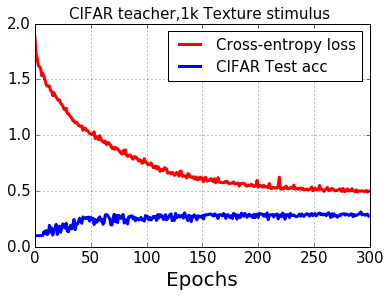}\\(a)&(b)\\
\end{tabular}
\caption{\label{fig:sv} Plot of cross entropy loss and test accuracy for each epoch.}
\end{figure}

Note that, the test accuracy and the cross entropy loss settles down with more iterations. Even with small training size (1k), overfitting is not observed. This could be because of soft labels used in the optimization.

\subsection{Quantification of Complexity}
We explored one of the quantification approach for complexity. We suspect that a stimulus dataset which matches the convolution filters as well as have more variations, work better for distillation. 
To validate this, we performed an experiment with CIFAR teacher and various 5k stimulus datasets. For the first convolution layer feature maps, we calculate a mean across the dataset. We also calculate an average of feature map-wise std. dev. for the dataset.  
A higher mean value indicates that convolution filters has better match with the dataset while high value of std.dev. indicates more variations.

\begin{table}[!h]
  \small
  \hspace*{2cm}\begin{tabular}{lllllll}
    \toprule
    \multirow{2}{*}{} & 
              
                 
      {SVHN} & {Shapes} & {Noise} & {MNIST} & {TinyImagenet} & {Texture}\\
      
      \midrule 
    Mean  & 0.117 & 0.134  & 0.139 & 0.148 & 0.165 & 0.174  \\
    Std Dev & 0.06 & 0.08  & 0.02 & 0.01 & 0.09 & 0.12  \\
    \bottomrule
  \end{tabular}
  \caption{\label{table:complexity} Mean and avg standard deviation for first layer filter map of CIFAR-teacher for various datasets. }
\end{table} 

We observe that a dataset with high value of mean as well as std. dev. works better for distillation.

\end{document}